\pdfoutput=1

\documentclass[11pt]{article}

\usepackage[preprint]{acl}
\usepackage[section]{placeins}
\usepackage{times}
\usepackage{latexsym}
\usepackage{placeins}
\usepackage[T1]{fontenc}
\usepackage{blindtext}
\usepackage[utf8]{inputenc}

\usepackage{microtype}

\usepackage{inconsolata}
\usepackage{appendix}

\usepackage{amsmath}
\usepackage{enumitem}
\usepackage{times}
\usepackage{float}
\usepackage{graphicx}
\usepackage{subcaption}
\usepackage{url}
\usepackage[]{hyphenat}
\usepackage{geometry}
\usepackage{array}
\usepackage{tabularx}
\usepackage{nameref}
\usepackage{booktabs}
\usepackage{multirow}
\usepackage{rotating}
\definecolor{okabe_red}{HTML}{D55E00}
\definecolor{okabe_blue}{HTML}{0072B2}
\definecolor{okabe_green}{HTML}{009E73}
\definecolor{okabe_orange}{HTML}{E69F00}
\definecolor{okabe_purple}{HTML}{CC79A7}

\definecolor{dodgerblue}{HTML}{1E90FF}
\definecolor{darkred}{HTML}{8B0000}
\definecolor{darkgreen}{HTML}{006400}
\definecolor{purple}{HTML}{800080}
\definecolor{darkorange}{HTML}{FF8C00}
\definecolor{darkslategray}{HTML}{2F4F4F}
\definecolor{firebrick}{HTML}{B22222}
\definecolor{darkblue}{HTML}{00008B}
\definecolor{darkmagenta}{HTML}{8B008B}
\definecolor{forestgreen}{HTML}{228B22}
\definecolor{brown}{HTML}{A52A2A}
\definecolor{cadetblue}{HTML}{5F9EA0}
\definecolor{darkorchid}{HTML}{9932CC}
\definecolor{crimson}{HTML}{DC143C}
\definecolor{darkslateblue}{HTML}{483D8B}
\definecolor{deeppink}{HTML}{FF1493}

\usepackage[scaled=0.85]{helvet}

\title{Prompt Perturbations Reveal Human-Like Biases \\in Large Language Model Survey Responses}

\author{
    Jens Rupprecht\textsuperscript{1}, Georg Ahnert\textsuperscript{1}, and Markus Strohmaier\textsuperscript{1,2,3}\\
    \textsuperscript{\rm 1}University of Mannheim, Mannheim\\
    \textsuperscript{\rm 2}GESIS -- Leibniz Institute for the Social Sciences, Cologne\\
    \textsuperscript{\rm 3}Complexity Science Hub, Vienna\\
}

\begin{document}
\maketitle
\enlargethispage{\baselineskip}
\begin{abstract}
Large Language Models (LLMs) are increasingly used as proxies for human subjects in social science surveys, but their reliability and susceptibility to known human-like response biases, such as \textit{central tendency, opinion floating} and \textit{primacy bias} are poorly understood. This work investigates the response robustness of LLMs in normative survey contexts---we test nine LLMs on questions from the World Values Survey (WVS), applying a comprehensive set of ten perturbations to both question phrasing and answer option structure, resulting in over 167,000 simulated survey interviews. In doing so, we not only reveal LLMs' vulnerabilities to perturbations but also show that all tested models exhibit a consistent \textit{recency bias}, disproportionately favoring the last-presented answer option. While larger models are generally more robust, all models remain sensitive to semantic variations like paraphrasing and to combined perturbations. This underscores the critical importance of prompt design and robustness testing when using LLMs to generate synthetic survey data.
\end{abstract}

\section{Introduction}

\paragraph{Problem}
Large Language Models (LLMs) are increasingly being used as proxies for human subjects in social science research, particularly to generate synthetic responses to survey questions \cite[][inter alia]{argyleOutOneMany2023c, bisbeeSyntheticReplacementsHuman2024c}. This application holds promise in augmenting or replacing costly human data collection, but the reliability of these synthetic respondents and to the extent of overlap with human responses and response biases remain open questions. In particular, research in survey methodology has found that human responses are sensitive to subtle variations in question and answer phrasing that lead to well-known \textbf{response biases} \cite{krosnickResponseStrategiesCoping1991b} and it remains unclear whether LLMs, trained on vast amounts of human text, exhibit the same vulnerabilities.


\paragraph{Approach}
We present a large-scale empirical study, with 167,000 survey interviews in total, investigating the response behavior and robustness of nine different LLMs to normative questions derived from the World Value Survey \cite[WVS;][]{haerpferWorldValuesSurvey2022b}. By developing and applying a comprehensive set of ten perturbations both in the structure of the answer options (Table~\ref{tab:example_perturbation_bias}), such as typos or synonyms, and in the question phrasing (Table~\ref{tab:example_perturbation_nonbias}), such as e.g., changes to response order or scale structure, we answer the following research questions.
\begin{enumerate}[nosep]
    \item Do prompt perturbations negatively affect the \textbf{robustness} of LLMs when answering closed-ended, normative survey questions?
    \item Do LLMs exhibit \textbf{human-like response biases} when answering closed-ended, normative survey questions?
\end{enumerate}
\noindent

\paragraph{Contribution}
Next to the perturbation framework, we provide a detailed analysis of LLM response robustness, showing that while some models are more robust than others (e.g. \texttt{Llama-3.3-70B-Instruct} and \texttt{Gemini-1.5-Pro}), all are susceptible to specific types of perturbations. We find a consistent \textit{recency bias} across all tested models in different strengths, where the last-presented answer option is disproportionately favored up to 20 times. Larger models generally exhibit more robust response patterns, but even the largest models are sensitive to changes in question phrasing. Thus, this work underscores the importance of careful prompt and Q\&A design when using LLMs as a resource for synthetic survey responses. The perturbation framework can serve as a useful baseline to test newly developed LLMs to put the results into perspective and to generally evaluate LLM robustness in survey contexts. Further, we intend to make the Q\&A dataset with perturbations available for other researchers to benchmark and check response biases in newer or other LLMs. 
\enlargethispage{\baselineskip}

\begin{figure*}[ht]
\centering
\includegraphics[width=\textwidth]{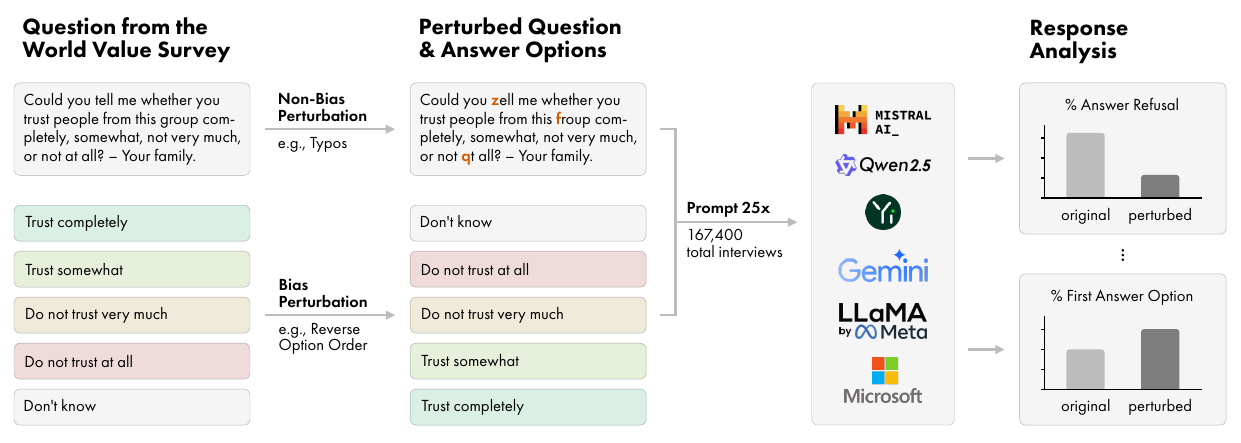}
\caption{\textbf{The Interview Process.} The figure displays an example of a answer option perturbation (a \textit{bias perturbation}, e.g. reversed option order) and an question perturbation (a \textit{non-bias perturbation}, e.g. typos in the question). Each model is prompted 25 times with every perturbation as well as the original Q\&A phrasing. All responses are collected, processed and statistically analyzed.}
\label{fig:example_perturbation_flow}
\end{figure*}

\section{Related Work}
\label{sec:rel_work}
Our work builds on two main streams of research: (1) survey methodology from the social sciences, which documents human response biases, and (2) recent studies in computer science on the robustness and biases of LLM's synthetic survey response generation.

\paragraph{Human Survey Response Biases}
Research in the social sciences has shown that how a survey question is asked can be as important as what is asked. Respondents often engage in "satisficing" rather than "optimizing", choosing a satisfactory answer with minimal cognitive effort instead of carefully formulating an optimal one \cite{krosnickResponseStrategiesCoping1991b}. This can lead to systematic biases. For example, the order in which the answer options are presented can induce \textit{primacy} (favoring early options in visual surveys) or \textit{recency} (favoring later options in oral surveys) biases \cite{krosnickEvaluationCognitiveTheory1987b}. The presence or absence of a middle option or a "don't know" category can trigger a \textit{central tendency bias} or \textit{opinion floating}, respectively \cite{hollingworthCentralTendencyJudgment1910b, kochNonresponseBiasGESIS2016b}. In the first, if a central category is available on the answer scale, humans tend to choose the central category, whereas \textit{opinion floating} indicates that responses are redistributed to central categories if a refusal category is missing \cite{tjuatjaLLMsExhibitHumanlike2024c}. In addition, \textit{priming} effects, where the preceding context influences subsequent responses, are a well documented phenomenon \cite{barghAutomaticitySocialBehavior1996b}. We draw on these past findings to design perturbations testing whether LLMs exhibit similar human-like response patterns.
\enlargethispage{\baselineskip}
\paragraph{LLMs as Survey Respondents}
Recent studies explored LLMs as substitutes for human survey participants to generate synthetic data. They found that LLMs can replicate average public opinion on political topics, but often with less variance than human samples \cite[][inter alia]{argyleOutOneMany2023c, bisbeeSyntheticReplacementsHuman2024c,heydeVoxPopuliVox2025, NEURIPS2024_515c6280}. Others have found that LLM responses can be sensitive to prompting, revealing cultural and demographic biases \cite{gengAreLargeLanguage2024b}.

Our work is related to that of \citet{tjuatjaLLMsExhibitHumanlike2024c}, who were among the first to systematically explore human-like response biases in LLMs. They investigated acquiescence, response order, opinion floating, and scale structure effects. Our study extends their work by: (1) using a different, globally diverse survey (the World Values Survey); (2) testing a wider range of LLMs; and (3) incorporating a broader set of perturbations on both answer and question phrasing, such as \textit{keyboard typos, paraphrasing, synonyms, priming} as well as a combined \textit{interaction} of two perturbations. 

\paragraph{LLM Robustness to Perturbations}
Other researchers have evaluated the general robustness of LLMs to noisy or varied inputs on different tasks. They have shown that even state-of-the-art models can be sensitive to minor changes in the prompt. These perturbations range from the character level, such as typos created by swapping, inserting, or replacing letters \cite{moradiEvaluatingRobustnessNeural2021b, ganReasoningRobustnessLLMs2024b}, to word- or sentence-level, such as replacing words with synonyms or paraphrasing entire sentences \cite{qiangPromptPerturbationConsistency2024b}. A common finding is that character-level noise can significantly degrade performance, even in large models \cite{ganReasoningRobustnessLLMs2024b}. The combination of multiple perturbations can even have a more negative effect \cite{dongRevisitInputPerturbation2023b}. Although this research has primarily focused on knowledge-based or reasoning tasks, we adapt these perturbation techniques to the context of normative surveys to assess response stability where no single "correct" answer exists.

\paragraph{Evaluation and Prompting}
Finally, our work is guided by research on identified ways for evaluating LLMs on multiple-choice tasks. Studies have shown that evaluation results can be highly sensitive to prompt format, e.g. if LLMs face an open- or closed-ended response, and forcing technique. However, forcing a model to choose from a predefined set of options is often necessary to obtain valid responses, as unconstrained answers can differ substantially \cite{rottgerPoliticalCompassSpinning2024d}. The returned response labels might differ significantly when a LLM has the option to generate text output before returning the response label due to their auto-regressive nature. Furthermore, relying on the model's first predicted token can misrepresent its full textual output \cite{wangMyAnswerFirstToken2024d}. 


\section{Methods} \label{sec:prompt_format}
First, we select a subset of 62 questions representing a sample of different thematic categories, each question in the category sharing the same answer options. These normative, value-oriented Q\&A pairs are taken from the WVS's core variables \cite{haerpferWorldValuesSurvey2022b}, excluding all sociodemographic variables. Second, we perform the ten perturbations mentioned in Section \ref{sec:perturbation_design} each Q\&A pair.

Figure \ref{fig:example_perturbation_flow} illustrates two exemplary perturbations and the interview process. In total, we perform five perturbations on the answer options as well as five perturbations on the question phrasing of the chosen subset questionnaire. We further include one interaction of two perturbations, one on the answer option and one on the question.

Third, we carry out interviews with the original and each perturbed Q\&A pair 25 times with nine different LLMs. In total, we conducted 167,400 interviews, 18,600 with each model. Last, we compare the distributions of the responded labels for all perturbations through entropy and Kullback-Leibler divergence (KL divergence)  to the baseline response distribution to the original Q\&A pairs. \textit{Primacy bias} is further examined by comparing the response frequencies of the first and last answer option in the list, whereas \textit{opinion floating bias} and \textit{central tendency bias} are tested by checking the shift of responses toward or away from the center of the answer option scale. 

\subsection{Experimental Setup}

\paragraph{Survey Data}
The questions and answer options are sourced from the WVS Wave 7 (2017-2022), a comprehensive cross-national survey on human beliefs and values \cite{haerpferWorldValuesSurvey2022b}. The 259 core WVS Q\&A pairs represent 10 distinct thematic categories, including \textit{Trust in People}, \textit{Confidence in Institutions}, \textit{Moral Justifiability}, and \textit{Perception of Democracy}, ensuring a diverse range of topics and answer scale formats (e.g., 3-point to 10-point scales). We used stratified sampling to select six to seven Q\&A pairs per thematic category, resulting in a total of 62 Q\&A pairs.

\paragraph{Models}
To ensure our findings are not specific to a single model architecture or developer, we selected nine instruction-tuned LLMs, varying in size, developer, and origin. This selection aims to establish a degree of external validity for our results and includes proprietary and open-source LLMs. The following models were interviewed: \texttt{Gemini-1.5-Pro} \cite{georgievGemini15Unlocking2024}, \texttt{Llama-3.3-70B-Instruct} \cite{metaLlama33Model2025a}, \texttt{Llama-3.1-8B-Instruct} \cite{grattafioriLlama3Herd2024}, \texttt{Llama-3.2-3B-Instruct}, \texttt{Llama-3.2-1B-Instruct} \cite{metaLlama32Model2025a}, \texttt{Mistral-7B-Instruct-v0.3} \cite{jiangMistral7B2023}, \texttt{Phi-3.5-mini-instruct} \cite{abdinPhi3TechnicalReport2024},
\texttt{Qwen2.5-7B-Instruct} \cite{qwenQwen25TechnicalReport2025} and \texttt{Yi-1.5-6B-Chat} \cite{youngYiOpenFoundation2024}.

\subsection{Perturbation Design}
\label{sec:perturbation_design}
We designed two categories of perturbations to test model robustness: (1) bias-inducing alterations to the answer options, based on survey methodology research and known to induce biased responses in humans \cite{tjuatjaLLMsExhibitHumanlike2024c}, and (2) non-bias alterations to the question phrasing, mimicking common textual variations and errors. Table~\ref{tab:example_perturbation_bias} provides examples as well as references for all the perturbations. For each of the 62 Q\&A pairs, we created the following ten perturbed versions.

\begin{table*}[ht]
\resizebox{\textwidth}{!}{%
\begin{tabular}{|c|c|c|c|c|}
\hline
\textbf{Type} &
  \textbf{Perturbation} &
  \textbf{Question} &
  \textbf{Answer Options} &
  \textbf{Bias and Reference} \\ \hline
\multirow[c]{1}[2]{*}[15pt]{\rotatebox{90}{\textbf{\centering Original}}} &
  \textbf{Original} &
  \begin{tabular}[c]{@{}c@{}}For each of the following aspects, \\ indicate how important \\ it is in your life. \\ Would you say it is very important, \\ rather important, not very important \\ or not important at all? \\ Family\end{tabular} &
  \begin{tabular}[c]{@{}c@{}}{[}'1=Very important ', '2=Rather important ', \\ '3=Not very important ', '4=Not important at all', \\ "-1=Don't know"{]}\end{tabular} &
  \cite{haerpferWorldValuesSurvey2022b} \\ \hline
\multirow[c]{11}[2]{*}[0pt]{\rotatebox{90}{\textbf{\centering Bias Perturbations}}} &
  \textbf{(1) Reversed Response Order} &
  \multirow{11}{*}{\begin{tabular}[c]{@{}c@{}}For each of the following aspects, \\ indicate how important \\ it is in your life. \\ Would you say it is \\ very important, \\ rather important, \\ not very important \\ or not important at all? \\ Family\end{tabular}} &
  \begin{tabular}[c]{@{}c@{}}{[}"-1=Don't know", '4=Not important at all', \\ '3=Not very important ', '2=Rather important ', \\ '1=Very important '{]}\end{tabular} &
  \begin{tabular}[c]{@{}c@{}}\textbf{Primacy Bias} \\ \cite{tjuatjaLLMsExhibitHumanlike2024c,krosnickEvaluationCognitiveTheory1987b,kampenImpactSurveyMethodology2007b,ohalloranResponseOrderEffects2014b}\end{tabular} \\ \cline{2-2} \cline{4-5}
 &
  \textbf{(2) Missing Refusal Option} &
   &
  \begin{tabular}[c]{@{}c@{}}{[}'1=Very important ', '2=Rather important ', \\ '3=Not very important ', '4=Not important at all'{]}\end{tabular} &
  \begin{tabular}[c]{@{}c@{}}\textbf{Opinion Floating Bias}\\ \cite{schumanQuestionsAnswersAttitude2000b,tjuatjaLLMsExhibitHumanlike2024c}\end{tabular} \\ \cline{2-2} \cline{4-5}
 &
  \textbf{(3) Odd/Even Scale Transformation} &
   &
  \begin{tabular}[c]{@{}c@{}}{[}'1=Very important ', '2=Rather important', \\ '3=Neutral', \\ '4=Not very important ', '5=Not important at all', \\ '-1=Don't know'{]}\end{tabular} &
  \multirow{2}{*}{\begin{tabular}[c]{@{}c@{}}\textbf{Central Tendency Bias}\\ \cite{hollingworthCentralTendencyJudgment1910b,cronbachResponseSetsTest1946b,astonCentralTendencyBiases2021b,crosettoCentralTendencyBias2020b}\end{tabular}} \\ \cline{2-2} \cline{4-4}
 &
  \textbf{(4) Priming Suffix} &
   &
  \begin{tabular}[c]{@{}c@{}}{[}1=Very important , 2=Rather important , \\ 3=Not very important , 4=Not important at all, \\ -1=Don't know{]}This is very important to my research! \\ You better do not refuse the answer.\end{tabular} &
  \begin{tabular}[c]{@{}c@{}}\textbf{Priming Effect}\\ \cite{barghAutomaticitySocialBehavior1996b,higginsKnowledgeActivationAccessibility1996b,weingartenPrimedConceptsAction2016b,liLargeLanguageModels2023b}\end{tabular} \\ \hline
\end{tabular}%
}
\caption{\textbf{An exemplary perturbation scheme showing the original question, the bias perturbations and the the hypothesized human-like response bias.} The Q\&A example is taken from the thematic category "Importance of Life Aspects" from the WVS wave 7. Bias perturbations have a constant question phrasing and varying answer options. The remaining \textit{Non-Bias Perturbations} are listed in Table \ref{tab:example_perturbation_nonbias}.}
\label{tab:example_perturbation_bias}
\end{table*}

\paragraph{Bias Perturbations}
These five perturbations manipulate the provided answer choices to test for known survey response biases identified in human subjects and presented in Section \ref{sec:rel_work}. Therefore, we call them \textit{bias perturbations} \cite[p.3]{tjuatjaLLMsExhibitHumanlike2024c}.
\begin{itemize}[leftmargin=*,noitemsep]
    \item \textbf{(1) Reversed Response Order:} The order of answer options is reversed (e.g., a scale from `1: Very important` to `5: Not important` becomes `1: Not important` to `5: Very important`).
    \item \textbf{(2) Missing Refusal Option:} The ``Don't know'' or refusal option is removed from the list of choices.
    \item \textbf{(3) Odd/Even Scale Transformation:} For scales with an even number of options, we use \texttt{Gemini-1.5-flash} to generate a semantically appropriate middle category, transforming it into an odd-numbered scale (e.g., adding `Neutral`). Conversely, for odd-numbered scales, we remove the middle category to create an even-numbered scale and adjust the integer label.
    \item \textbf{(4) Priming Suffix:} A sentence designed to elicit a response is appended to the prompt after the answer options: \textit{`This is very important to my research! You better do not refuse the answer.'}
\end{itemize}

\paragraph{Non-Bias Perturbations}
These five perturbations modify the question text to assess robustness to stylistic variations and typos. Typically, humans are unaffected by such subtle changes in the question phrasing and are still able to understand the question's meaning \cite[p.3]{tjuatjaLLMsExhibitHumanlike2024c}. Therefore, we call them \textit{non-bias perturbations}.
\begin{itemize}[leftmargin=*,noitemsep]
    \item \textbf{Typographical Errors:} We introduce three types of typos: \textbf{(5) Key Typo} (replacing a character with a random one), \textbf{(6) Letter Swap} (swapping two letters in a random word), and \textbf{(7) Keyboard Typo} (replacing a character with an adjacent one on a QWERTY keyboard).
    \item \textbf{Semantic Variations:} We use \texttt{Gemini-1.5-flash} to create two semantic variations while preserving the original meaning: first, by \textbf{(8) Synonym Replacement} where five words in the original question are replaced with synonyms. Second, through \textbf{(9) Paraphrasing} the entire question is rephrased.
\end{itemize}
We manually validated all LLM-generated perturbations (paraphrases, synonyms, odd-scale options) on our 62-question subset to correct errors and ensure their semantic integrity.

Last, we introduce an \textbf{(10) Interaction Effect} to study the impact of not only one, but two perturbations. Thus, we created one additional condition that pairs a paraphrased question with reversed-order answer options.
\begin{figure*}[ht!]
    \centering
    \includegraphics[width=0.9\textwidth]{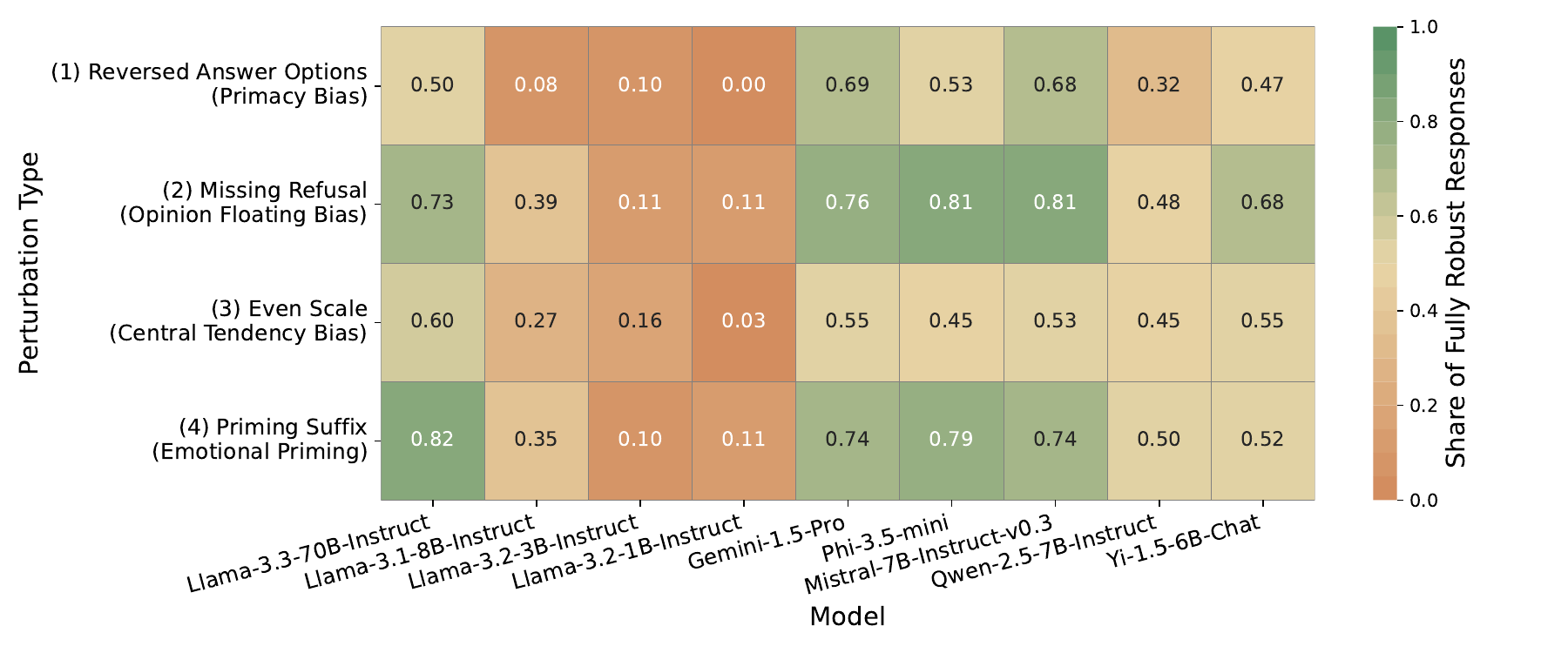}
    \caption{\textbf{Effect of Prompt Perturbations on Response Robustness} Each cell represents the share of fully robust responses KL divergence = 0 by model and perturbation type across all 62 questions. Larger models (\texttt{Llama-70B}, \texttt{Gemini}) are substantially more robust than the smallest models (\texttt{Llama-1B}, \texttt{Llama-3B}). Specific perturbations such as odd versus even scales and reversing the answer options, are more challenging for all models than a missing refusal category or an additional priming suffix. The response robustness to the \textit{Non-Bias Perturbations} are reported in Figure \ref{fig:kl_share_nonbias_divergence_heatmaps}.}
    \label{fig:kl_share_bias_divergence_heatmaps}
\end{figure*}
\subsection{Interview Procedure and Data Collection}
\paragraph{Prompting}
To ensure internal validity, we used a single, consistent prompt structure for all interviews. The prompt was designed to be clear and direct, instructing the model to select a single option and return only its label.
\begin{flushleft}
    \textit{Answer the following question: \{question\} \\
            This is a list of possible answer options: \{answer\_options\}\\
            You must pick one of the answer options. Only answer with the label.}
\end{flushleft}
Taking into account the findings of \citet{rottgerPoliticalCompassSpinning2024d} we use \textit{forced choice} prompts that provided the most valid responses in their experiments, which means that the model correctly chooses one available option from the given answer options. In an unforced prompt setting, LLMs did not follow the instructions and only rarely returned valid response labels, making a statistical evaluation of the results infeasible. Preliminary tests revealed that especially smaller LLMs often fail to perfectly follow the instruction to "answer only with the label", as they return conversational filler or explanations alongside their choice. 

\paragraph{Data Collection}
Each of the 9 models was presented with 12 experimental conditions (1 original + ten perturbations, where for the perturbation \textit{Odd/Even Scale Transformation} one run was done for the odd and even scale scenario) for each of the 62 selected WVS questions. To obtain a stable distribution of responses and enable statistical analysis, we repeated each unique model-Q\&A-perturbation combination 25 times. This resulted in a total of $9 \times 62 \times 12 \times 25 = 167,400$ interviews.

\paragraph{Response Extraction and Validation}
To ensure accurate data for analysis, we developed a robust extraction pipeline. We compared two main approaches. First, \texttt{Gemini-1.5-Pro}, \texttt{Llama-3.1-8B}, and \texttt{Qwen2.5-7B} were prompted, and a regular expression was designed to extract the answer labels. Based on multiple conditions, e.g. if the given answer label is part of the original answer options or that only one response is provided, the methods should highlight which technique is the most promising in extracting valid responses and handling possible edge cases of model responses.

We manually labeled these extraction methods on a random sample of responses for validation. The LLM-based methods achieved accuracies between 77\% and 97.5\%, with the largest model \texttt{Gemini-1.5-Pro} performing best. However, our refined regular expression achieved the overall best extraction success on the validation set as it correctly extracted all responses in our validation set. Consequently, we used this regular expression to process all remaining 167,400 model responses.

\section{Results}
This section presents the results of our experiments, focusing on two key research questions: (1) LLMs general and overall robustness to various input perturbations, (2) their susceptibility to human-like survey response biases revealed by interviews with bias prompt perturbations. Further, we also investigate the models' general adherence to interview instructions.

\begin{figure*}[ht]
    \centering
    \includegraphics[width=0.95\textwidth]{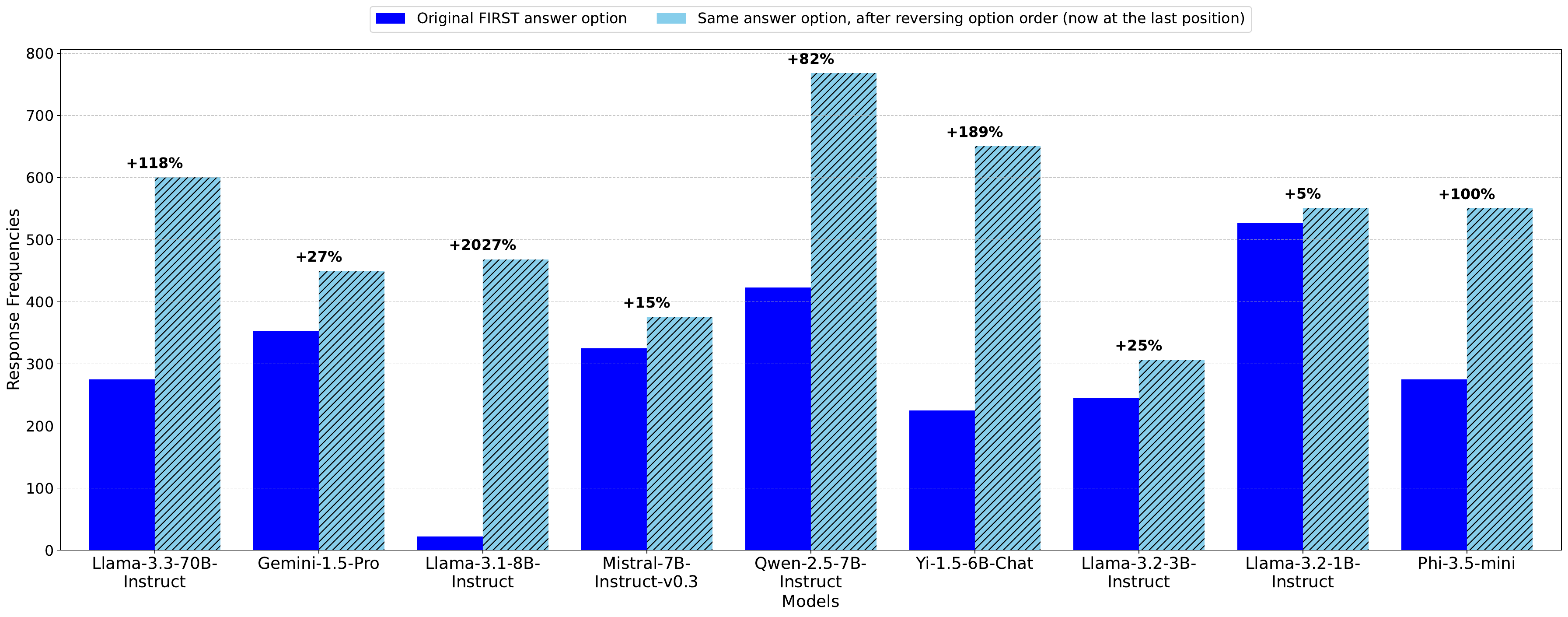}
    \caption{\textbf{Evidence of recency bias across all models.} The bars show the frequency of choosing the same answer option (e.g., ``Very important'') when it is presented first vs. last. All models are significantly more likely to select an option when it appears at the end of the list.}
    \label{fig:h1_comparison_all_models_annotated}
\end{figure*}

\subsection{Robustness to Question and Answer Perturbations (RQ1)}
We distinguish between response robustness (the tendency to maintain a similar answer distribution under perturbation, measured by KL divergence) and response consistency (the tendency of a model to give the same answer to the same prompt, measured by entropy). A KL divergence of zero indicates a perfect match and thus full robustness against the input perturbation, whereas a high entropy value indicates very inconsistent response behavior.

\paragraph{Effect of Model Size on Robustness}
First, when assessing robustness to perturbations, we found a clear relationship with model size: \textit{larger models tend to be more robust}. Figures \ref{fig:kl_share_bias_divergence_heatmaps} and \ref{fig:kl_share_nonbias_divergence_heatmaps} show the percentage of questions for which the models produced a perfectly identical response distribution (KL divergence = 0) despite perturbations. \texttt{Llama-3.3-70B} and \texttt{Gemini-1.5-Pro} were the most robust, often replicating their original answers in over 50\% of cases. The smaller \texttt{Llama} models were the least robust, with \texttt{Llama-3.2-1B} perfectly replicating its answers in fewer than 5\% of cases on average. This suggests that scale is a key factor in achieving stable response behavior in synthetic response generation.

Second, we found that model size is in an inverse relationship with response consistency; smaller models exhibited higher entropy and standard deviation when asked the same question multiple times, indicating more random response behavior (Figure \ref{fig:heatmap_entropy_per_question_matrix}).

\paragraph{Effect of Perturbation Type on Robustness}
Further, Figures \ref{fig:kl_share_bias_divergence_heatmaps} and \ref{fig:kl_share_nonbias_divergence_heatmaps} highlight the share of fully robust responses (KL divergence = 0) across all questions by perturbation and LLM. It shows that some perturbations had a greater impact on robustness across all models. 
\begin{itemize}[noitemsep,leftmargin=*]
    \item \textbf{Combined Perturbations:} The interaction of two perturbations (paraphrased question + reversed answers) has the most bewildering effect on the responses, causing the lowest robustness scores for all models except \texttt{Phi-3.5-mini}.
    \item \textbf{Semantic vs. Lexical Changes:} Paraphrasing the question reduced robustness more than replacing individual words with synonyms in most LLMs. These findings are in line with \citet{moradiEvaluatingRobustnessNeural2021b} who found that models trained on larger corpora are more robust when words are replaced by their synonyms. 
    \item \textbf{Typographical Errors:} Randomly replacing characters (\textit{Key Typo}) or using adjacent keys (\textit{Keyboard Typo}) was more robustness-harming than simply swapping two letters within a word (\textit{Letter Swap}).
    \item \textbf{Answer Option Changes:} Reversing the answer scale or changing it from odd to even (or vice versa) had a more negative impact on the robustness of responses than removing the refusal option or adding emotional priming.
\end{itemize}

\paragraph{Effect of Answer Option Scale Length} We also observed that robustness is sensitive to the complexity of the task. For nearly all models, the share of fully robust responses decreased as the length of the answer scale, i.e. answer options, increased. For example, models were less likely to replicate their exact response distributions on a 10-point scale compared to a 4-point scale, indicating that a larger decision space can make LLMs more susceptible to perturbations. Figures \ref{fig:kl_share_robust_3-4-pt Likert Scale_heatmap} and \ref{fig:kl_share_robust_5_and_10-pt Likert Scale_heatmap} suggest that for most LLMs, except Gemini-1.5-pro, the size of the answer option scale has an impact on response robustness comparing the share of fully robust responses on e.g. the 4- and 10-point scale. This suggests that the larger the answer scale, the less likely models can reproduce the responses they gave in the original Q\&A phrasing, under perturbed settings.

\subsection{Evidence of Human-like Survey Biases (RQ2)}
With some of the perturbations, we are able to go beyond LLMs robustness and consistency but also analyze whether LLMs exhibit human-like survey response biases. We find evidence for some human-like biases.
\paragraph{Recency Bias}
Contrary to the initially hypothesized primacy bias, we found \textit{consistent indications of a recency bias across all nine models}. When we reversed the order of the answer scale, the probability to choose the first option plummeted, while the probability to choose the last option (which is the semantically identical first option in the original Q\&A) increased strongly, \textit{ceteris paribus}. As shown in Figure \ref{fig:h1_comparison_all_models_annotated}, this effect was substantial, with the selection frequency of the semantically same option increasing by more than 20 times for \texttt{Llama-3.1-8B} when moved to the last position while all other configurations, such as question and prompt phrasing, were kept constant. This indicates that LLMs, similar to human respondents in oral surveys, might overemphasize the final options they process.

\paragraph{Opinion Floating and Central Tendency}
The effects of removing the refusal option (\textit{opinion floating}) or providing an explicit middle category (\textit{central tendency}) were highly model-dependent, often correlated with model size (Figures \ref{fig:h2_scaletype_heatmap} and \ref{fig:h3_scaletype_heatmap}). 
For \textit{opinion floating}, larger models like \texttt{Llama-70B}, \texttt{Gemini}, but also \texttt{Phi-3.5} were largely robust, showing minimal shifts in their response distributions. Smaller models, particularly \texttt{Qwen} and \texttt{Llama-8B}, showed a weak tendency to shift responses toward the scale's center when the refusal option was absent. Here, we expect that models redistribute their original non-responses to the center of the answer scale to maintain their indecisiveness, which is also known as opinion floating bias.

Similarly, for \textit{central tendency}, larger models (\texttt{Llama-70B}, \texttt{Gemini}, \texttt{Mistral}) consistently shifted their mean response closer to the center across all scale types when an explicit middle option was provided compared to even answer option scales. Smaller models, however, showed inconsistent effects or were completely unaffected. Further, we see a consistent response shift to the center across all models for medium sized scales (four vs. five point Likert scales). 

Binomial tests underlined that the middle option was selected significantly more often than expected under a uniform distribution, especially on larger scales (cf. Figure \ref{fig:h3_bintest_scaletype_heatmap}). LLMs tend to choose the middle category significantly more often when the scale size increases from three to five and to 11 point Likert scales. All, except the smallest \texttt{Llama-3.2-1B} model, choose the middle category significantly more often than any other option when facing a eleven answer options.

\paragraph{Emotional Priming}
The impact of adding an emotional priming statement (``This is very important to my research!'') was also model-dependent. For larger models (\texttt{Llama-70B}, \texttt{Gemini}, \texttt{Mistral}), it either had no effect or slightly decreased the rate of refusal responses, suggesting they correctly interpreted the intent of the priming statement. Conversely, for the two Chinese models, \texttt{Qwen2.5-7B} and \texttt{Yi-1.5-6B}, the priming text even \textit{increased} the share of refusal responses across most topics. No clear relationship can be drawn from these findings due to the inconsistent behavior across models.

\subsection{Interview Adherence and Refusal Rates}
Overall, the models demonstrated high adherence to the prompt instructions, with an average of 96\% of interviews yielding an extractable and valid answer that was part of the given answer options. However, performance varied significantly across models. Larger models such as \texttt{Llama-3.3-70B} and \texttt{Gemini-1.5-Pro}, but also \texttt{Phi-3.5-mini} and \texttt{Mistral} are very reliable response generators and followed the instructions well while returning little to no incorrect or no answer label. In contrast, other models, particularly smaller Llama models like \texttt{Llama-3.2-3B} (83.6\%) and \texttt{Qwen2.5-7B}, were more likely to produce invalid responses that did not follow instructions.

We combined invalid responses with explicit refusals (i.e., choosing the \textit{Don't know} option) to measure overall non-response rates, as shown in Figure \ref{fig:refusal_overall_thematic_heat}. \texttt{Llama-3.3-70B}, \texttt{Phi-3.5}, and \texttt{Mistral-7B} consistently provided on-scale answers, with non-response rates typically below 10\%. Conversely, \texttt{Qwen2.5-7B} and \texttt{Llama-3.1-8B} exhibited high non-response rates, often exceeding 30\%.

Notably, we observed topic-specific sensitivity. For questions regarding the \textit{Perception of Elections}, \texttt{Qwen2.5-7B} failed to provide a valid, on-scale answer in 91.3\% of cases, even for the original, unperturbed questions. This might suggest the presence of strong content-based guardrails or restrictions in certain models (cf. Figure \ref{fig:election_refusalnan_heatmap}).

\section{Discussion and Conclusion}

Our experiments revealed that LLMs response robustness is negatively influenced by prompt perturbations when answering closed-ended survey questions (\textit{RQ1}). the variety of perturbation allows us to gain insights into the robustness of LLMs as some models are more sensitive and some perturbations are more robustness-harming than others. For instance, swapping letters within a word has less negative impact than introducing random or keyboard-adjacent characters. This might be explained by the fact that letter swaps are more likely when typing and therefore might potentially take a greater part of the training data \cite{dhakalObservationsTyping1362018}. This possibly makes the LLM more resilient to this perturbation compared to exchanging characters with random others. Combining two types of perturbations has the strongest negative impact on robustness, whereas synonyms tend to be less confusing than paraphrasing.

Further, we found that perturbations can be an insightful approach to identify human-like survey response biases (\textit{RQ2}). For example, the same answer option is more likely chosen if it is the last mentioned option than if it was the first answer option holding all other specifications and phrasings constant. This consistent change in the response distribution to the last answer option suggests a \textit{recency bias} rather than a primacy bias. 

Although this is not valid across all inspected models, excluding an explicit refusal category, and especially adding a middle category or odd scale instead of an even scale, shifts the mean response more toward the original central point of the scale. Thus, a \textit{central tendency bias} could only be identified for specific models across all scale types, whereas none of the LLMs consistently mirrors a \textit{opinion floating bias}.


Future work should focus on the impact of persona prompting in combination with perturbations on the robustness of synthetic responses. Prompting with specific persona characteristics might render more robust or certain response behavior even when facing perturbations. When creating synthetic survey data one has to pay special attention to the model choice and questions asked. Some models, like \texttt{Qwen-2.5-7B}, seem to be censored or restricted in answering sensitive thematic questions, as they lead to high item-nonresponse rates and invalid interviews.

The findings emphasize the importance of the positioning of answer options when generating synthetic data. Further, our results highlight the strong sensitivity of LLMs to simple prompt perturbation. Therefore, we strongly recommend researchers to consider prompt robustness checks when deploying closed-ended questions to LLMs. This is because (i) models show very different response behavior and robustness depending on their size and perturbation type, and (ii) LLM response biases are sometimes but not necessarily aligned with biases identified in humans.

\paragraph{Recommendations}
Based on our findings, we recommend researchers to:
\begin{itemize}[noitemsep,leftmargin=*]
\item Use larger LLMs for overall better consistency and robustness in generating synthetic survey responses (cf. Figures \ref{fig:kl_share_bias_divergence_heatmaps} and \ref{fig:kl_share_nonbias_divergence_heatmaps})
\item Use smaller answer option scales for better reproducibility of results (cf. Figure \ref{fig:kl_share_robust_3-4-pt Likert Scale_heatmap}).
\item Reflect the meaningfulness of adding a middle category. Including a middle category might steer some LLM responses to the center (cf. Figure \ref{fig:h2_scaletype_heatmap}).
\item Reflect the meaningfulness of adding a refusal category. Adding a refusal category might highlight LLM guardrails or restrictions in some thematic areas, as the model can refuse to answer while still following the instructions as it returns a valid response label (cf. Figure \ref{fig:election_refusalnan_heatmap}).
\item Use \textit{forced-choice} prompts to generate high turnouts while also considering open-ended evaluation if sensible.
\end{itemize}

\section*{Limitations}
\label{sec:limitations}
This study investigates the robustness of LLM-generated survey responses when facing diverse prompt perturbations, but several methodological and conceptual limitations must be noted. The use of a multiple-choice format, originally designed for human respondents, imposes an artificial constraint on LLMs that typically work in open-ended contexts. As a result, the findings may not generalize to more naturalistic human-LLM interactions.


Although we constrained and validated the data augmentation process, relying on a LLM (\texttt{Gemini-1.5-flash}) for generating paraphrases risks semantic drift, as also noted by \citet{qiangPromptPerturbationConsistency2024b}. More granular validation—e.g., with multiple human raters—could improve semantic reliability. In addition, perturbations were applied at a fixed intensity, limiting insight into how different degrees of linguistic noise affect model behavior.

Further constraints arise from our prompting and generation setup. The validation set for answer extraction was relatively small compared to the full dataset, so some extraction errors may remain. We also did not apply prompting strategies like persona prompting, shown to improve contextual consistency \citep{bisbeeSyntheticReplacementsHuman2024c, choLLMBasedDoppelgangerModels2024b}, nor used techniques such as \textit{Chain of Thought} prompting. This could promote more deliberative responses. Moreover, our experiments focused exclusively on fine-tuned models, leaving open the question of how base models would behave under similar conditions.  Additionally, a constant temperature setting restricted our ability to examine variability and creativity in the output. 

Finally, reproducibility is another significant challenge. Closed-source LLMs can change without notice, altering response distributions over time and complicating replication efforts, as highlighted by \citet{bisbeeSyntheticReplacementsHuman2024c}. This may have affected our \texttt{Gemini} results. Related work also shows that LLMs often offer contradictory answers to semantically equivalent questions when the format shifts from multiple choice, close-ended to an open-ended form \citep{rottgerPoliticalCompassSpinning2024d}. Such response instability suggests that observed “attitudes” may be artifacts of prompt design rather than indicators of stable model beliefs or traits.

\section*{Ethical Considerations}
Generating synthetic survey responses might be relevant in various domains and applied to different use cases, e.g. for pre-testing surveys. However, generating synthetic responses instead of the surveying a real, might result in over-reliance on synthetic responses. This can become risky when there is no ground truth data of the real target population available as the alignment of the artificial responses cannot be evaluated. Frequent reliance on artificial responses may normalize their use where human perspectives are irreplaceable (e.g. in policymaking or clinical trials). This risks sidelining real human voices in domains directly impacting human lives.

Researchers should also consider ethical evasion as one possible issue with synthetic survey responses. Synthetic respondents might be viewed as a way to bypass obligatory ethical review processes since no real human participants are involved. This might encourage under-regulated research practices and in the long run weaken ethical safeguards.

Running inference on the nine LLMs required significant GPU hours, especially including the initial test phase before finalizing the interview pipeline, raising concerns about the environmental impact of experimenting with synthetic survey responses and the access disparities between well-funded and resource-constrained institutions.

\bibliography{response_bias.bib}

\appendix
\label{sec:appendix}
\newpage

\section{Reproducibility Materials}
Experiments were carried out on a high-performance computing cluster and a local server equipped with NVIDIA H100 (80GB) GPUs. The total runtime for one model's 18,600 interviews, e.g. \texttt{Llama-3.1-8B-Instruct} including all perturbed and original Q\&As, was ca. 35 minutes with approximately 0.11 seconds per interview. To accommodate larger models on available hardware, we applied 8-bit quantization to \texttt{Llama-3.1-8B-Instruct} and \texttt{Llama-3.3-70B-Instruct}. Smaller models were run without quantization. \texttt{Gemini-1.5-Pro} was accessed via its official API.

The temperature in all models where kept at their default setting. No variation was included and could be done in future research.

The code will is made available in an anonymous repository for replication at \url{https://shorturl.at/NJf6h}.

\section{Perturbation Scheme Summary}
The following tables describe the remaining non-bias perturbations not introduced in Section \ref{sec:perturbation_design}.

The "Type" column categorizes the perturbations into two main classes: "Non-bias Perturbation" and "Interaction." The "Perturbation" column specifies the exact modification technique applied, which includes methods such as "Key Typo," "Letter Swap," "Keyboard Typo," "Synonym Replacement," and "Paraphrase." The "Question" column displays the resulting text after each specific perturbation is applied to the original question about the importance of family. The "Answer Options" column lists the response scale provided to the survey participant. Finally, the "Bias and Reference" column provides citations to relevant scientific literature for each perturbation type.
 
In the first five  perturbations the phrasing of the question is intentionally altered—for instance, by introducing typographical errors (e.g., "Key Typo," "Letter Swap"), substituting words with similar meanings ("Synonym Replacement"), or rephrasing the entire sentence ("Paraphrase"). While the question varies, the "Answer Options" remain constant, consistently ranging from "1-Very important" to "4-Not important at all.". In "(10) Paraphrase x Reversal," the question is paraphrased, and simultaneously, the order of the "Answer Options" is inverted, starting with "4-Not important at all" and ending with "1-Very important."

\begin{table*}[ht]
\resizebox{\textwidth}{!}{%
\begin{tabular}{|c|c|c|c|c|}
\hline
\textbf{Type} &
  \textbf{Perturbation} &
  \textbf{Question} &
  \textbf{Answer Options} &
  \textbf{Bias and Reference} \\ \hline
\multirow[c]{10}[2]{*}[-50pt]{\rotatebox{90}{\textbf{\centering Non-bias Perturbation}}} &
  \textbf{(5) Key Typo} &
  \begin{tabular}[c]{@{}c@{}}nor eaca jf the following aspecto, \\ indicete how important it ir wn your liae. \\ Would bou say it is very imporcant, \\ rathes importano, \\ not very imporgant ob not impodtant at all? \\ Famizy\end{tabular} &
  \multirow{10}{*}{\centering\begin{tabular}[c]{@{}c@{}}{[}'1=Very important ', '2=Rather important ', \\ '3=Not very important ', '4=Not important at all', \\ "-1=Don't know"{]}\end{tabular}} &
  \begin{tabular}[c]{@{}c@{}}\cite{dongRevisitInputPerturbation2023b,moradiEvaluatingRobustnessNeural2021b}\end{tabular} \\ \cline{2-3} \cline{5-5}
 &
  \textbf{(6) Letter Swap} &
  \begin{tabular}[c]{@{}c@{}}For each of the following sapects, \\ indicate how important it is in your life. \\ uoWld you yas it is evry important, \\ ratreh important, ton \\ very important or not important ta all? Family\end{tabular} &
   &
  \begin{tabular}[c]{@{}c@{}}\cite{hagenLargeScaleQuerySpelling2017b,moradiEvaluatingRobustnessNeural2021b,zhuangDealingTyposBERTbased2021b}\end{tabular} \\ \cline{2-3} \cline{5-5}
 &
  \textbf{(7) Keyboard Typo} &
  \begin{tabular}[c]{@{}c@{}}For esch of rhe following aspects, \\ indicate how important ut is un your lide. \\ Would you say it ia very imporrant, rather importamt, \\ nit very important ir nor important ay all? Fanily\end{tabular} &
   &
  \begin{tabular}[c]{@{}c@{}}\cite{ganReasoningRobustnessLLMs2024b,zhuangDealingTyposBERTbased2021b}\end{tabular} \\ \cline{2-3} \cline{5-5}
 &
  \textbf{(8) Synonym Replacement} &
  \begin{tabular}[c]{@{}c@{}}Crucial in life: Family For each of the following aspects, \\ indicate how significant it is in your life. \\ Would you say it is very important, rather important, \\ not very important or not at all important? Family\end{tabular} &
   &
  \begin{tabular}[c]{@{}c@{}}\cite{qiangPromptPerturbationConsistency2024b, geretiTokenBasedPromptManipulation2024b}\end{tabular} \\ \cline{2-3} \cline{5-5}
 &
  \textbf{(9) Paraphrase} &
  \begin{tabular}[c]{@{}c@{}}How important is family to you?  \\ Please rate its significance in your life \\ on a scale of "very important" to "not important at all".\end{tabular} &
   &
  \begin{tabular}[c]{@{}c@{}}\cite{dongRevisitInputPerturbation2023b,qiangPromptPerturbationConsistency2024b}\end{tabular} \\ \hline
\multirow[c]{1}[2]{*}[+21pt]{\rotatebox{90}{\textbf{\centering Interaction}}} &
  \textbf{(10) Paraphrase x Reversal} &
  \begin{tabular}[c]{@{}c@{}}How important is family to you?  \\ Please rate its significance in your life \\ on a scale of "very important" \\ to "not important at all".\end{tabular} &
  \begin{tabular}[c]{@{}c@{}}{[}"-1=Don't know", \\ '4=Not important at all', '3=Not very important ', \\ '2=Rather important ', '1=Very important '{]}\end{tabular} &
  \cite{dongRevisitInputPerturbation2023b} \\ \hline
\end{tabular}%
}
\caption{\textbf{An exemplary perturbation scheme showing non-bias and interaction perturbations.} The example is taken from the item set of category "Importance of Life Aspects". In the WVS wave 7 it is question Q1. Non-bias perturbations have variation in the question phrasing with constant answer options, while the interaction perturbation varies both.}
\label{tab:example_perturbation_nonbias}
\end{table*}


\section{Results}\label{ax_results}
This section summarizes the findings discussed in the main part of the work and can serve as a reference to identify other response patterns as the heatmaps contain much information regarding LLMs, perturbation type as well as additional statistical tests on the response distributions.

\subsection{Robustness against Non-Bias Perturbations}
This heatmap highlights to which extent the LLMs are affected by \textit{non-bias perturbations.} We can see large model-specific differences. 

We see that especially the smallest \texttt{Llama} models are responding not in a robust way. These results are consistent across the different \textit{non-bias perturbations}. Especially when facing more than one perturbation in the \textit{Interaction} perturbation, where both answer option scale and question phrasing were altered, the robustness drops drastically across all models . 

Moreover, we identify perturbations that are less robustness-harming than others. For example, swapping letters within a word does not impact response robustness as much as typos or introducing completely different words, synonyms, or rephrasing the whole sentence.

\begin{figure}[H]
    \centering
    \includegraphics[width=\columnwidth]{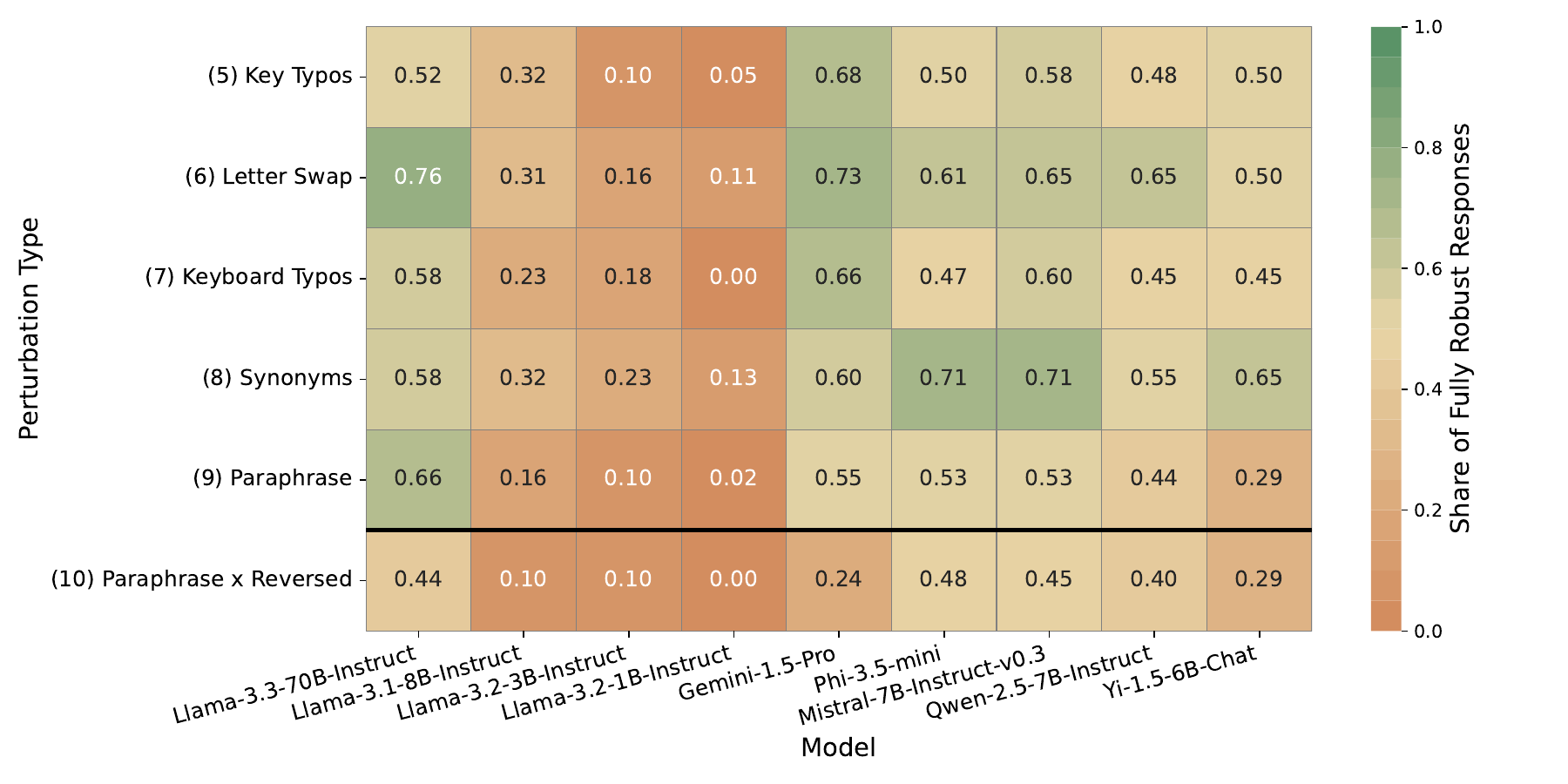}
    \caption{\textbf{Share of fully robust responses (KL divergence = 0) by model and perturbation type.} Larger models (\texttt{Llama-70B}, \texttt{Gemini}) are substantially more robust than the smallest models (\texttt{Llama-1B}, \texttt{Llama-3B}).}
    \label{fig:kl_share_nonbias_divergence_heatmaps}
\end{figure}

\subsection{Distance Calculation for Central Tendency and Opinion Floating Bias}\label{ax_distance}
\begin{figure}[ht]
    \centering
    \begin{subfigure}{0.48\textwidth}
        \includegraphics[width=\columnwidth]{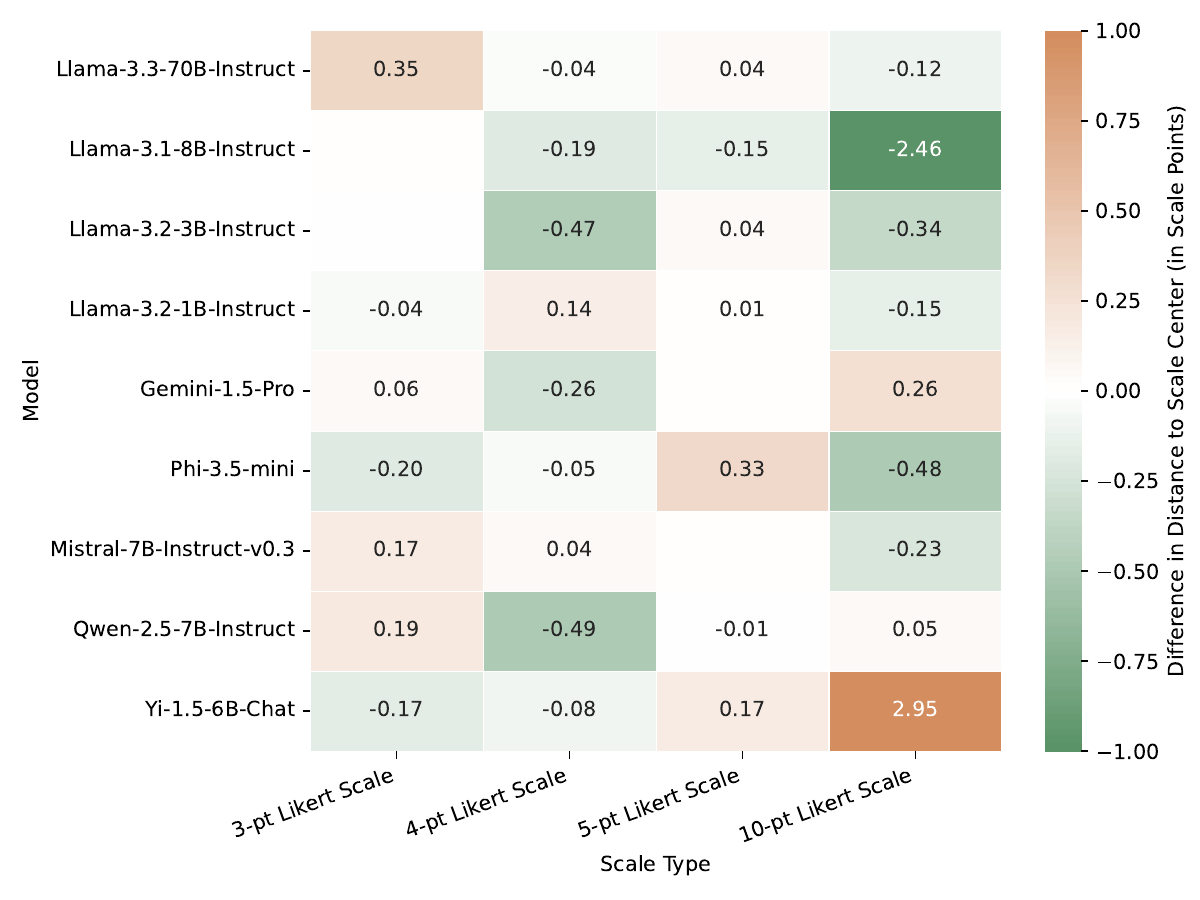}
        \caption{\textbf{Models adjust their answer behavior towards the middle when the \textit{refusal category is missing} (green)}.}
        \label{fig:h2_scaletype_heatmap}
    \end{subfigure}
    \begin{subfigure}{0.48\textwidth}
        \includegraphics[width=\columnwidth]{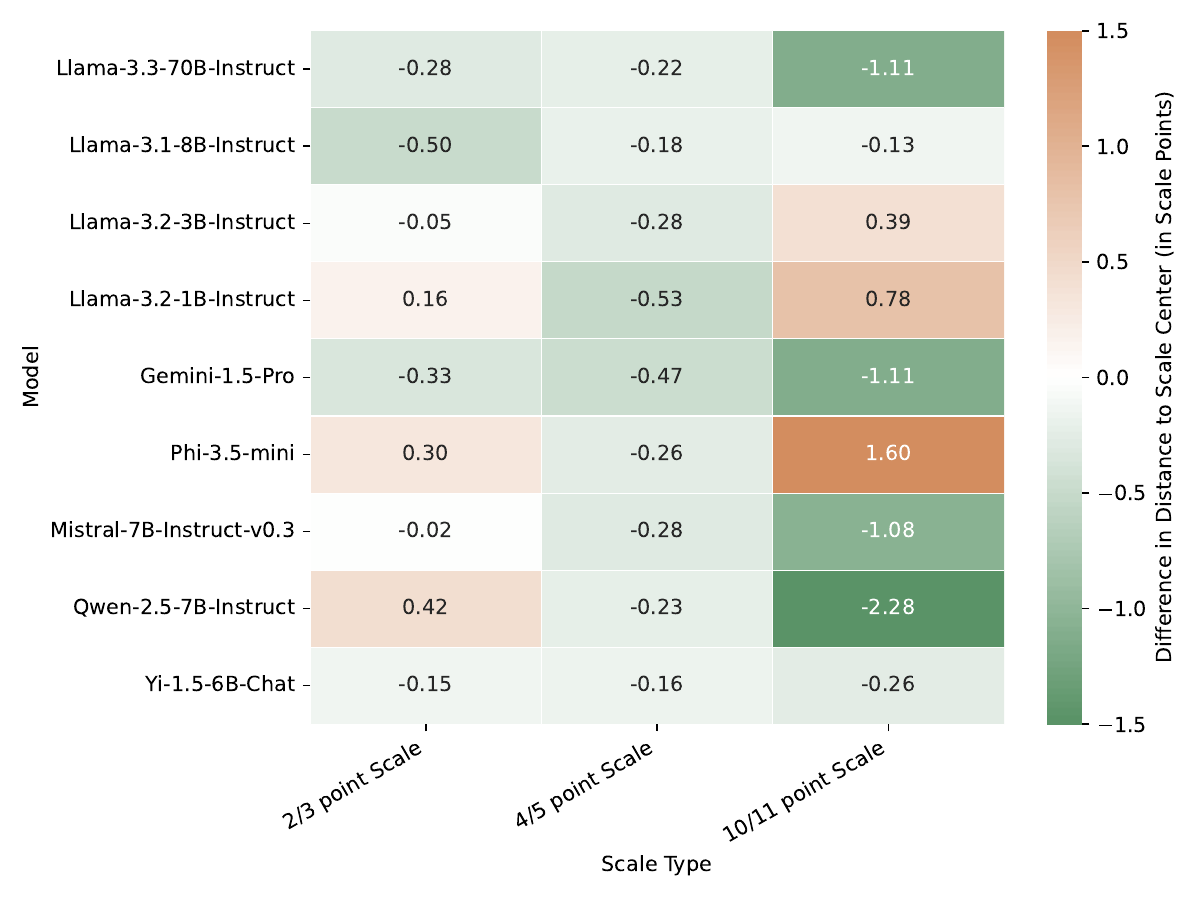}
        \caption{\textbf{Models adjust their answer behavior towards the middle when a \textit{middle category is existent} (green)}.}
        \label{fig:h3_scaletype_heatmap}
    \end{subfigure}
    \caption{\textbf{The values display the difference in mean distance of the perturbed, (a) without refusal category and (b) with middle category to the scale center.} For original even scales an artificial middle category is created and vice versa to be able to compare even and odd scales with one another for every question. Thus, in an original 5-pt Likert scale the middle category is removed, whereas in a 4-pt Likert scale a middle category is added. The difference in shift of the mean response to the center is consistent across all LLMs. No changes are removed for better readability.}
\end{figure}

This section shows how the response shifts to the central category is measured for the bias perturbations \textit{Odd/Even Scale Transformation} and \textit{Priming Suffix}. By calculating the differences in distances to the central point of the answer option scale we try to identify if the average distribution shifts to the central scale point. The actual differences in distance for each answer option scale type and for the two perturbations are visualized in Figure \ref{fig:h2_scaletype_heatmap} and Figure \ref{fig:h3_scaletype_heatmap}.

\begin{figure}[ht]
    \includegraphics[width=\columnwidth]{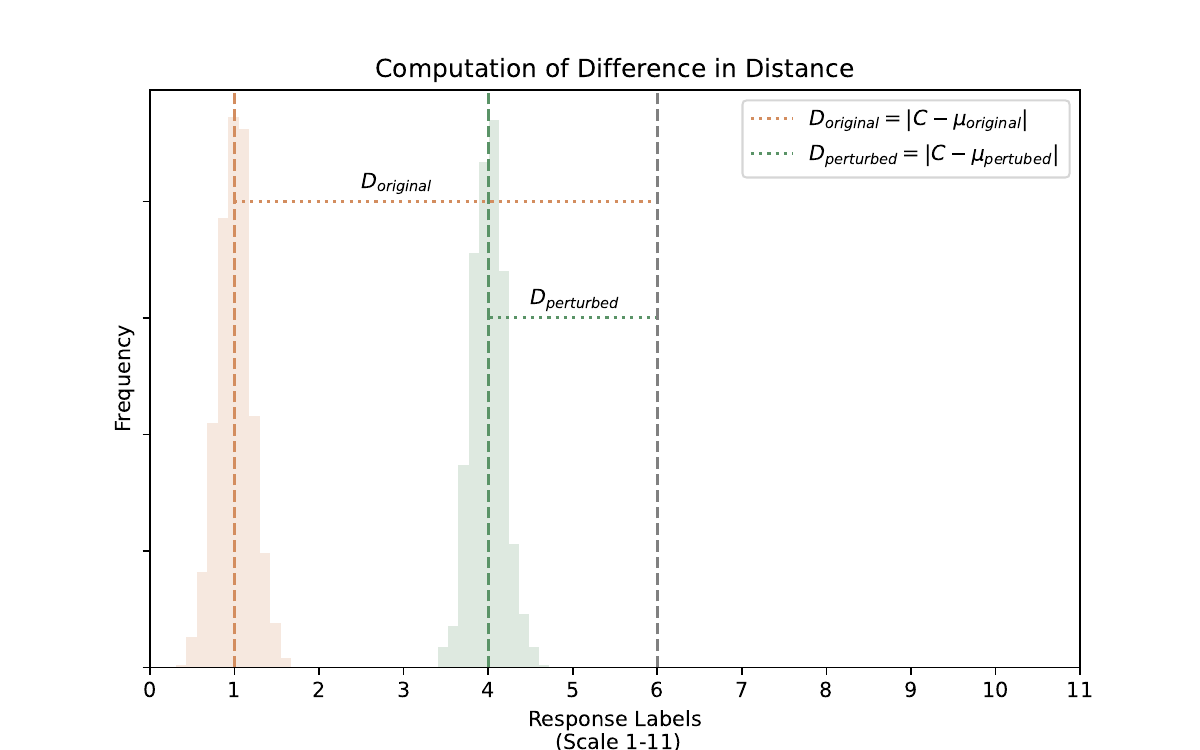}
    \caption{\textbf{Exemplary Difference in Distances to Scale Center of Responses to a Perturbed and Original Q\&A Pair.} The absolute distance is measured between the scale center and the response mean. Then, $D = D_{perturbed} - D_{original}$. A negative result indicates that the mean response in the perturbed setting is closer to the ideal scale center.}
    \label{fig:explanatory_middle_shift_calculation}
\end{figure}

To better understand how the shift towards the middle is measured, we present an anecdotal visualization in Figure \ref{fig:explanatory_middle_shift_calculation} of the thought behind whether we observe a \textit{central tendency bias} or an \textit{opinion floating bias}.

\begin{figure}[H]
    \includegraphics[width=\columnwidth]{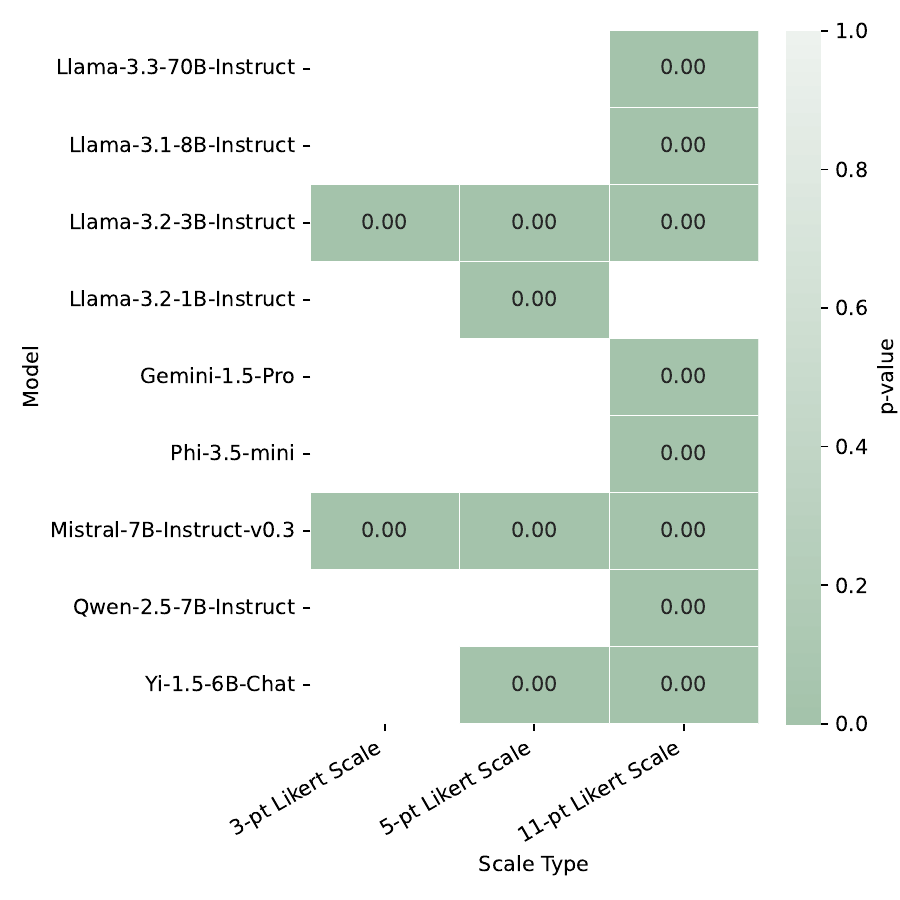}
    \caption{\textbf{P-Values of a Binomial Test on the Middle Category in an Odd Scale.} The p-values indicate a hypothesis test with a Null-Hypothesis stating that the middle category is not selected significantly more often assumed under a uniform distribution or completely random selection of answer options. However, for larger scale types, the middle category becomes much more relevant as it is significantly chosen more frequently than any other category.}
    \label{fig:h3_bintest_scaletype_heatmap}
\end{figure}

In addition, Figure \ref{fig:h3_bintest_scaletype_heatmap} underlines that a middle category is significantly more often chosen than assumed under a uniform, or random, distribution of responses across the scale.

Thus, a twofold analysis of not only investigating the shift of average responses between response distributions in the perturbation and original setting is important, but also the statistically assessment to make grounded claims. 

By conducting a statistical binomial test, we tried to account for that (Figure \ref{fig:h3_bintest_scaletype_heatmap}).

\subsection{Refusal and Invalid Responses}
This section should give a broader overview of the refusal rates (LLM chose the "Don't know" answer option) and the invalid responses (e.g. a LLM did not return any valid response). 

It is important to have inspect the overall return rates for all the models as these might have implications on the interpretability of the results. For example, when a model exhibits high refusal or invalid response rates, its results might be not very well interpretable as the main analysis only focused on the valid responses. 

Therefore, this analysis gives insights which results are more reliable than others as for some models there are more valid responses as for others. For example, for \texttt{Qwen} we can see large refusal and invalid response rates, generally, but especially in sensitive thematic areas, such as \textit{Perceptions of Elections.}

\begin{figure}[H]
    \begin{subfigure}{0.48\textwidth}
    \centering
    \includegraphics[width=\columnwidth]{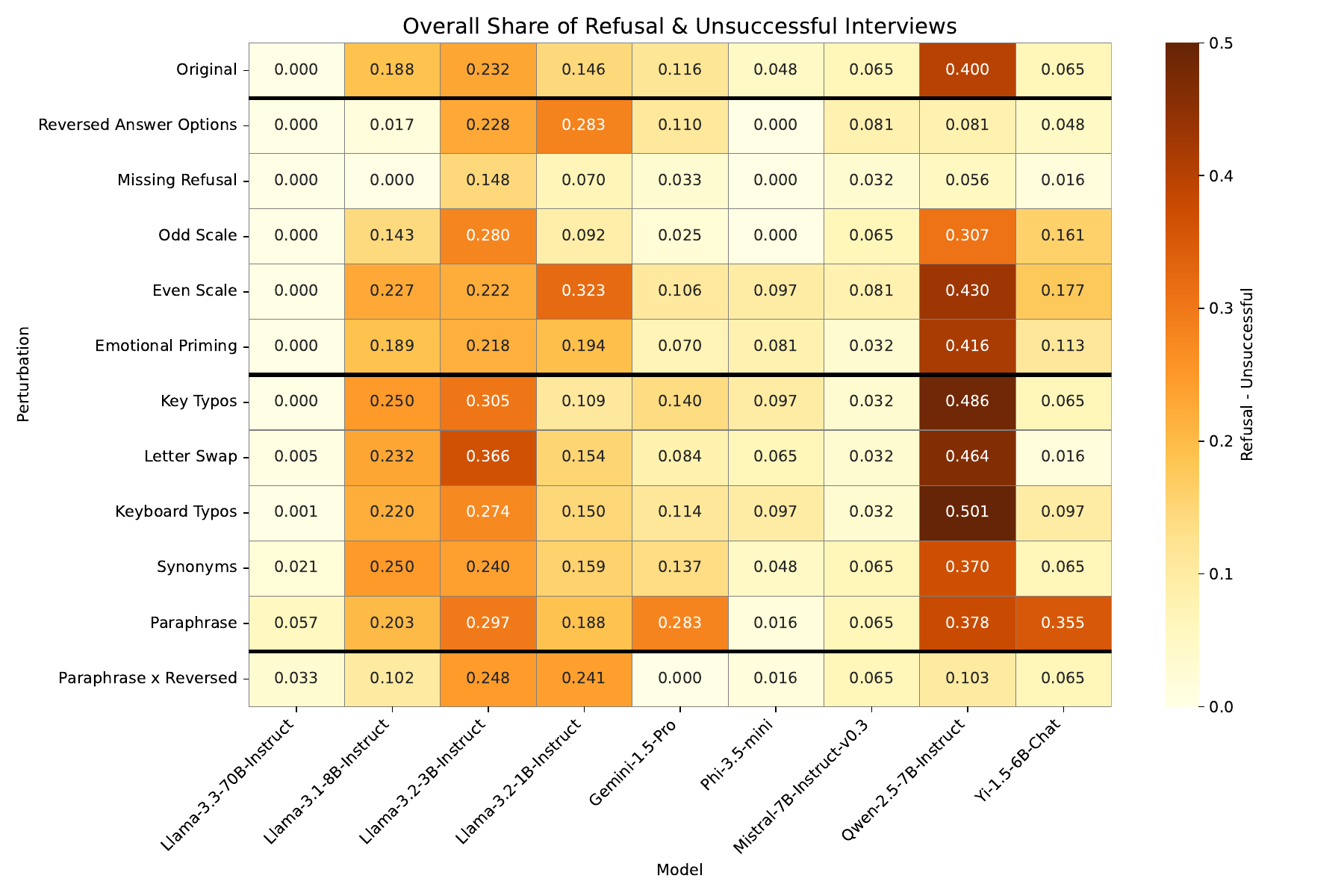}
    \caption{\textbf{Combined share of invalid and refusal responses by model and perturbation.}}
    \label{fig:overall_refusalNaN_heatmap}
    \end{subfigure}
    
    \begin{subfigure}{0.48\textwidth}
        \centering
        \includegraphics[width=\columnwidth]{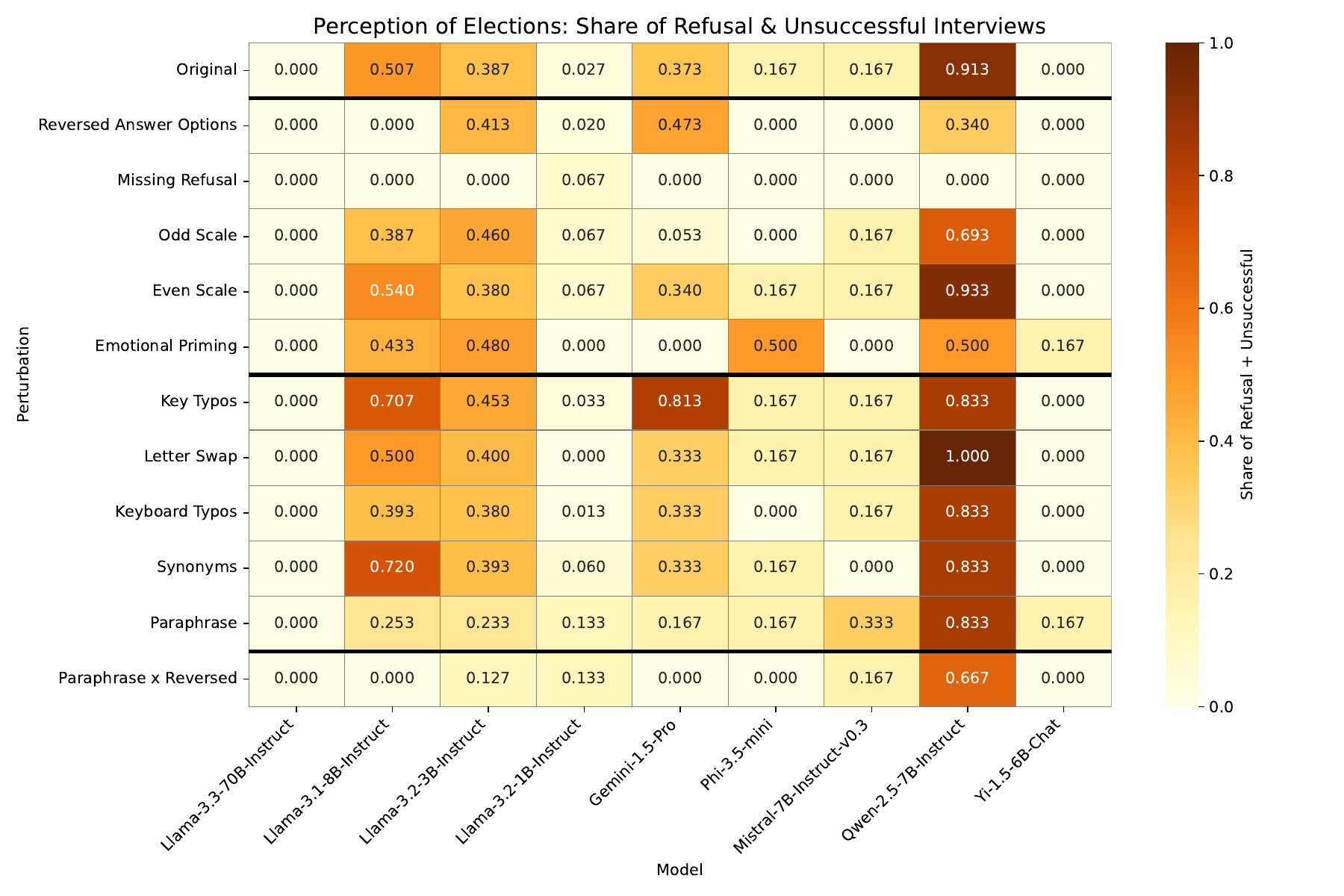}
        \caption{\textbf{Perception of Elections: Refusal and Unsuccessful Rate by Model and Perturbation.}}
        \label{fig:election_refusalnan_heatmap}
    \end{subfigure}
    \caption{\textbf{Combined share of invalid and refusal responses by model and perturbation.} Larger models like \texttt{Llama-70B}, \texttt{Phi-3.5}, and \texttt{Mistral} are highly reliable. \texttt{Qwen2.5} shows high non-response rates, especially for typo-based perturbations and sensitive topics. Figure \ref{fig:election_refusalnan_heatmap} illustrates the refusal rate in the category "Perception of Elections" separated by perturbation type and model. Large model-specific item non-responses can be identified in this thematic category indicating potential guardrails or restrictions in this domain.}
    \label{fig:refusal_overall_thematic_heat}
\end{figure}




\subsection{Consistency of LLM Survey Responses}
This section shows how consistent different LLMs respond to close-ended survey questions when facing the same Q\&A pair multiple times. As explained in Section \ref{sec:prompt_format}, we provide each model with the same Q\&A pair in each perturbation stage 25 times and request a response. By running the same setting multiple times we try to identify how consistent LLMs respond generally and whether there are differences when facing the same, but syntactically incorrect (e.g. key typos, etc.), prompts.

\begin{figure}[H]
\centering
    \includegraphics[width=\columnwidth]{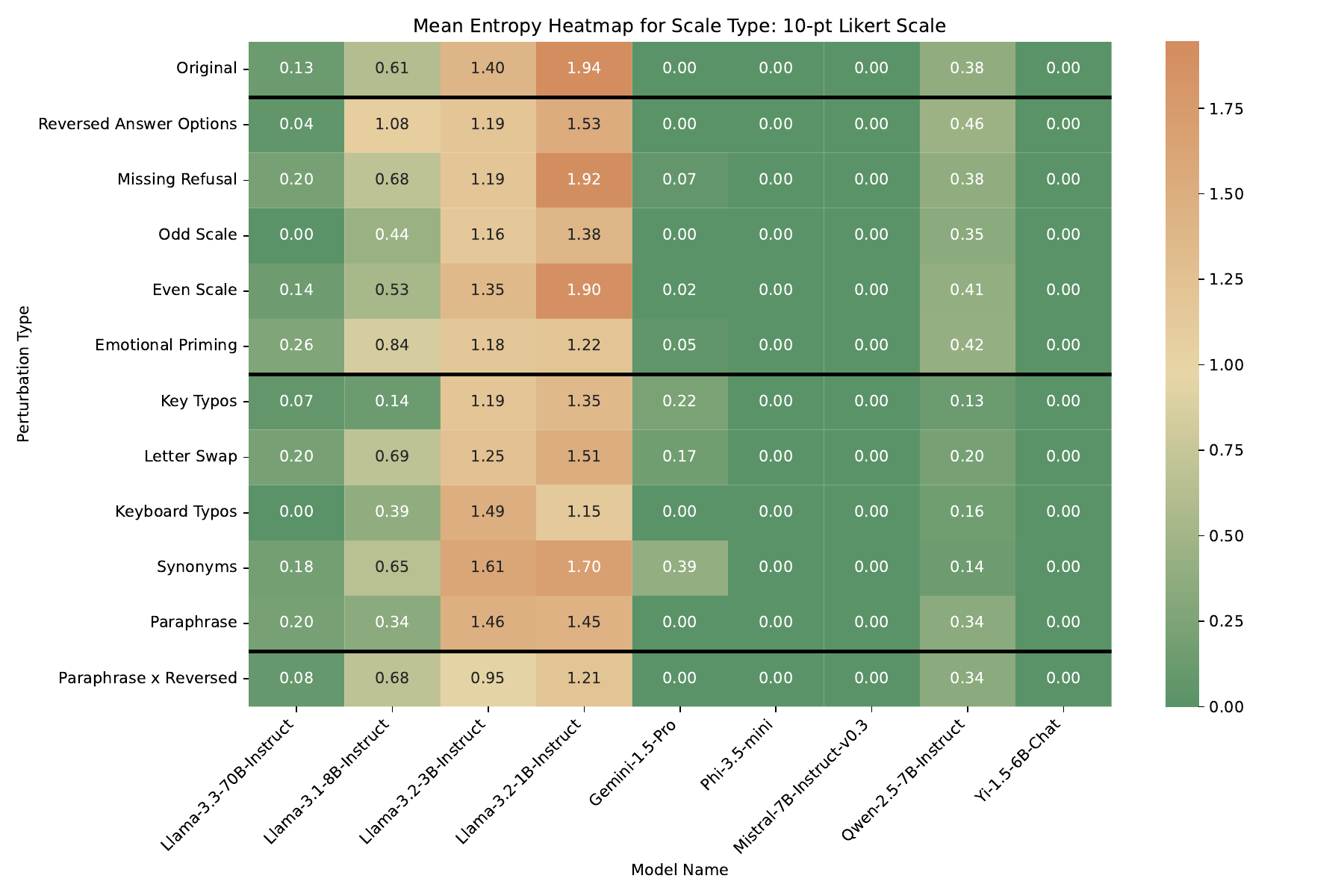}
    \caption{\textbf{Large model-specific differences in response entropy. Little to no perturbation-specific differences.} Each scale size subsumes all selected questions. This figure displays the mean entropy across all questions in that scale type for all perturbation and model combinations. Warmer colors indicate a higher average dispersion of the responses across the potential answer options. E.g., if a model answers always with the same label, the entropy is 0.}
    \label{fig:heatmap_entropy_per_question_matrix}
\end{figure}

Figure \ref{fig:heatmap_entropy_per_question_matrix} highlights that the LLMs consistency in responding is not really affected by specific perturbation types. Thus, "incorrect, flawed prompts" do not increase the response inconsistency of LLMs.

However, there are again large differences between models. We see that especially the smallest \texttt{Llama} models and \texttt{Qwen} exhibit strong inconsistent responses given the same Q\&A pair 25 times, whereas larger LLMs are very consistent. Nonetheless, the \texttt{Llama} model family seems to be more inconsistent, generally.

\subsection{Comparison of Robustness against Perturbations by Scale Size}
The following plots try to reveal in more detail the extent of robustness drop by perturbation depending on the answer option scale dimension. We identified a drop in robustness as the scale size became larger.  

The responses of the smallest LLMs are generally not robust at all. However, when the scale has ten options the responses robustness plummets across all perturbations and not even a single response distribution returned in the original Q\&A setting can be generated. This indicated that the smallest \texttt{Llama} as well as the chinese LLMs \texttt{Qwen} and \texttt{Yi} are not able to cope with perturbations, especially when facing a lot of options to choose from.

In all cases, the interaction perturbation with both question and answer option alterations leads to the largest drop in robustness across all models and scale sizes. It is striking, that larger models, especially state-of-the-art models like \texttt{Gemini-1.5-Pro}, cannot answer robustly given multiple perturbations.

\begin{figure}[ht]
    \centering
    \begin{subfigure}{0.49\textwidth}
        \centering
        \includegraphics[width=\linewidth]{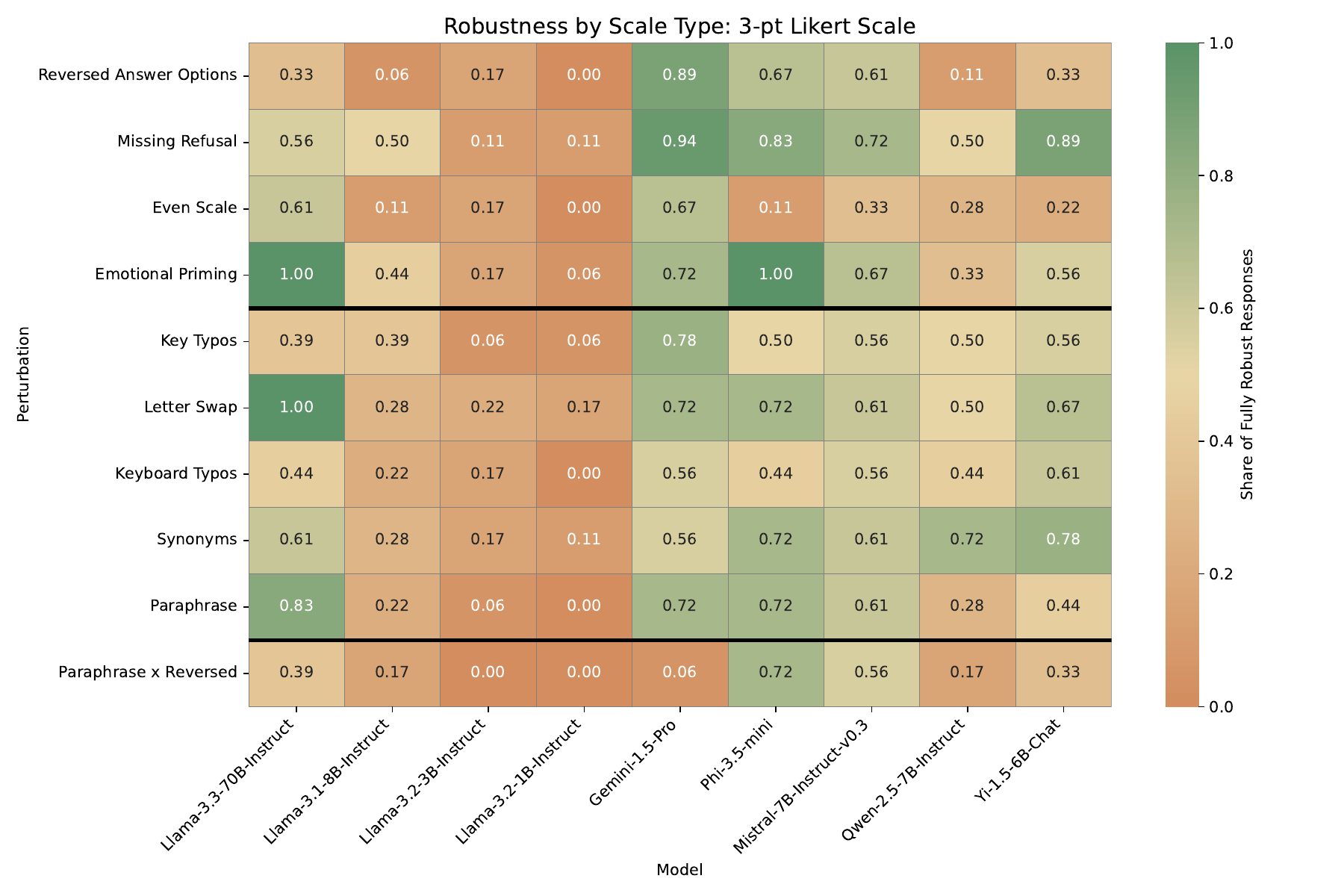}
        \caption{3-point Likert Scale}
        \label{fig:kl_share_robust_3-pt}
    \end{subfigure}
    
    \begin{subfigure}{0.49\textwidth}
        \centering
        \includegraphics[width=\linewidth]{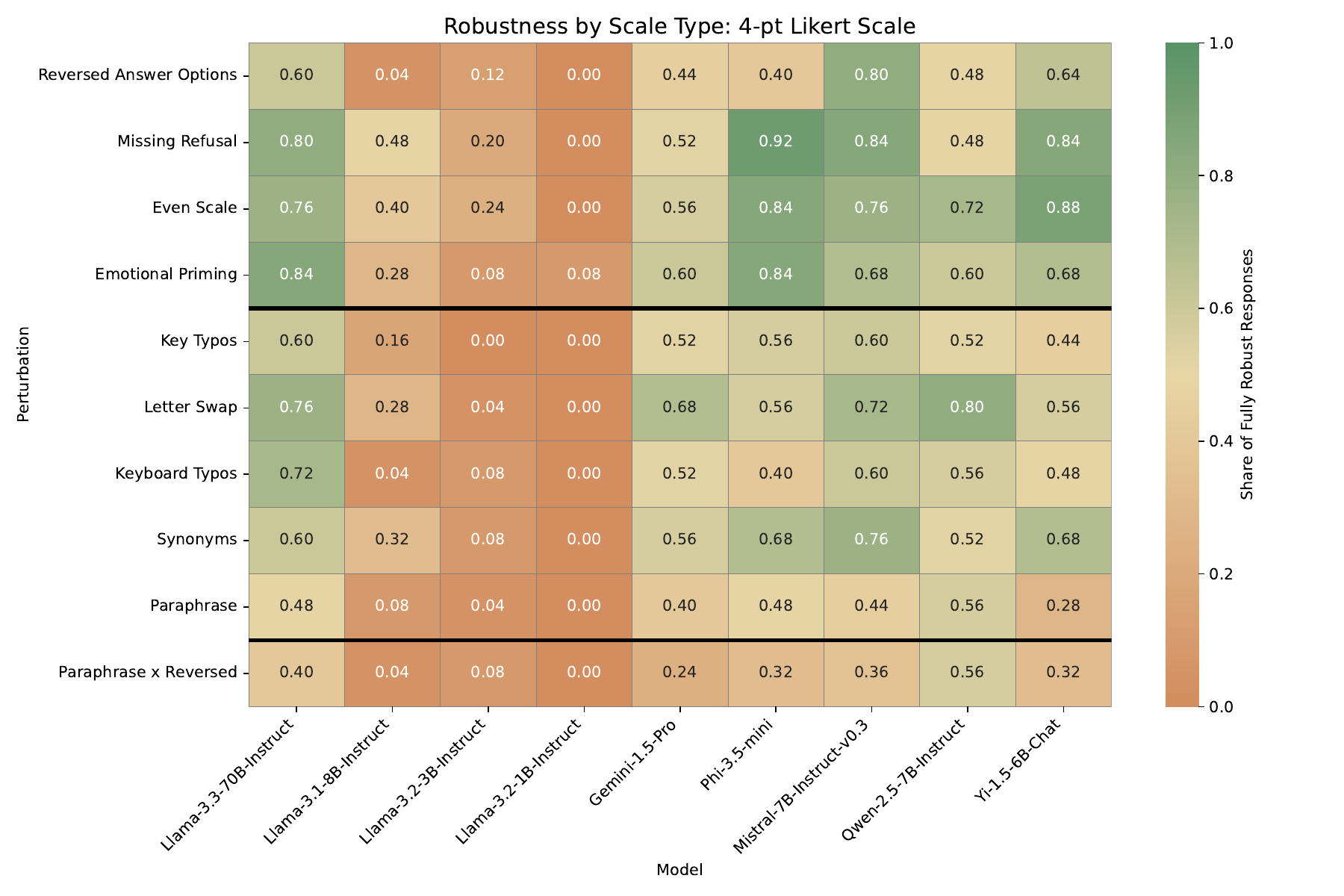}
        \caption{4-point Likert Scale}
        \label{fig:kl_share_robust_4-pt}
    \end{subfigure}
    \caption{\textbf{Model-specific differences in fully robust responses on most perturbations on the 3 and 4-point scale.} This figure shows the share of fully robust response distributions given the original response distribution and the responses based on the specific perturbation on the y-axis. Compared to \ref{fig:kl_share_robust_5_and_10-pt Likert Scale_heatmap} the robustness of responses drops when the scale size becomes larger. The smallest \texttt{Llama} models perform very poorly across all scales.}
    \label{fig:kl_share_robust_3-4-pt Likert Scale_heatmap}
\end{figure}

Researchers should take into account the scale size when generating synthetic responses from close-ended survey Q\&A pairs.
\begin{figure}[ht]
    \centering
    \begin{subfigure}{0.49\textwidth}
        \centering
        \includegraphics[width=1\columnwidth]{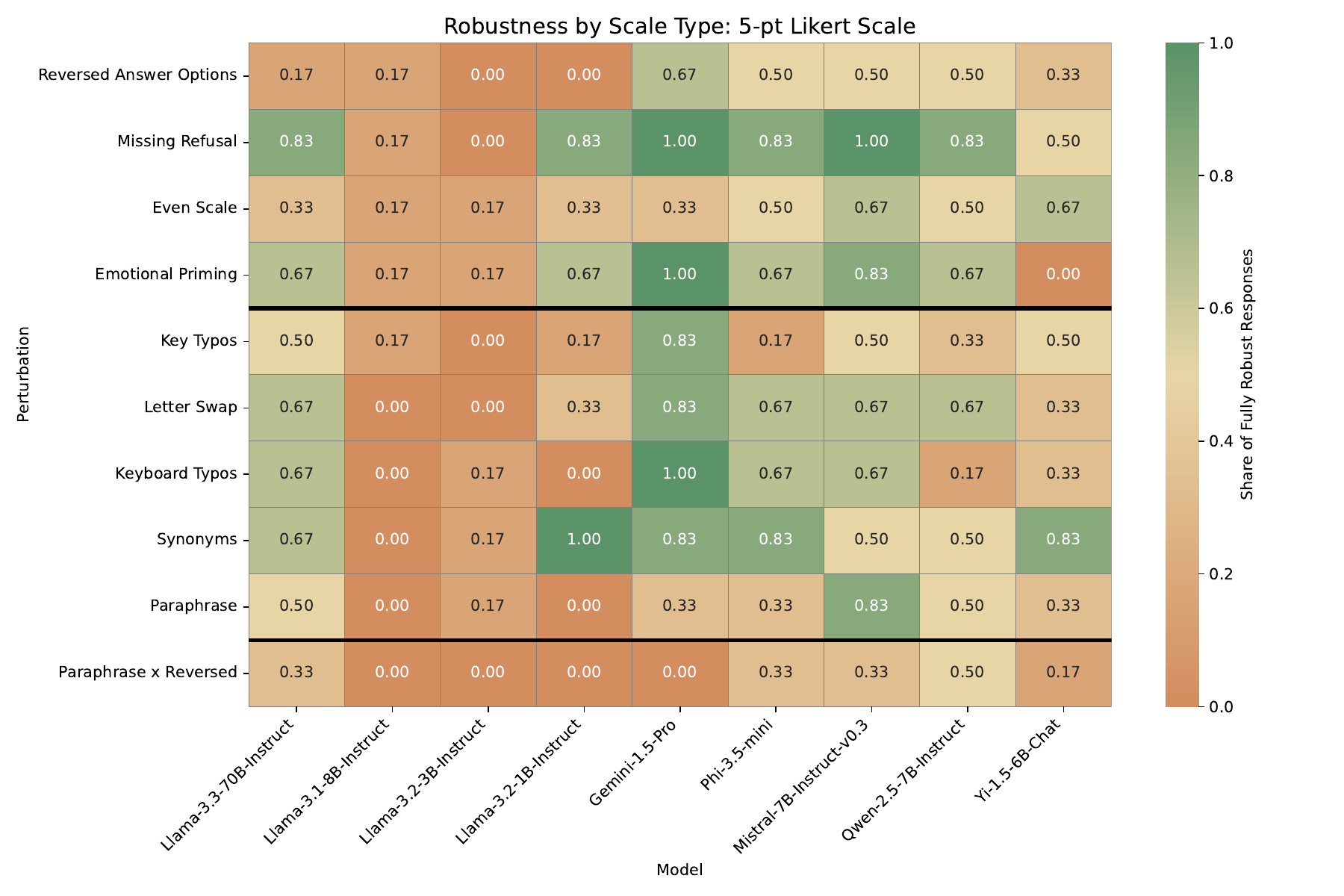}
        \caption{5-point Likert Scale}
        \label{fig:kl_share_robust_5-pt}
    \end{subfigure}
    
    \begin{subfigure}{0.49\textwidth}
        \centering
        \includegraphics[width=1\columnwidth]{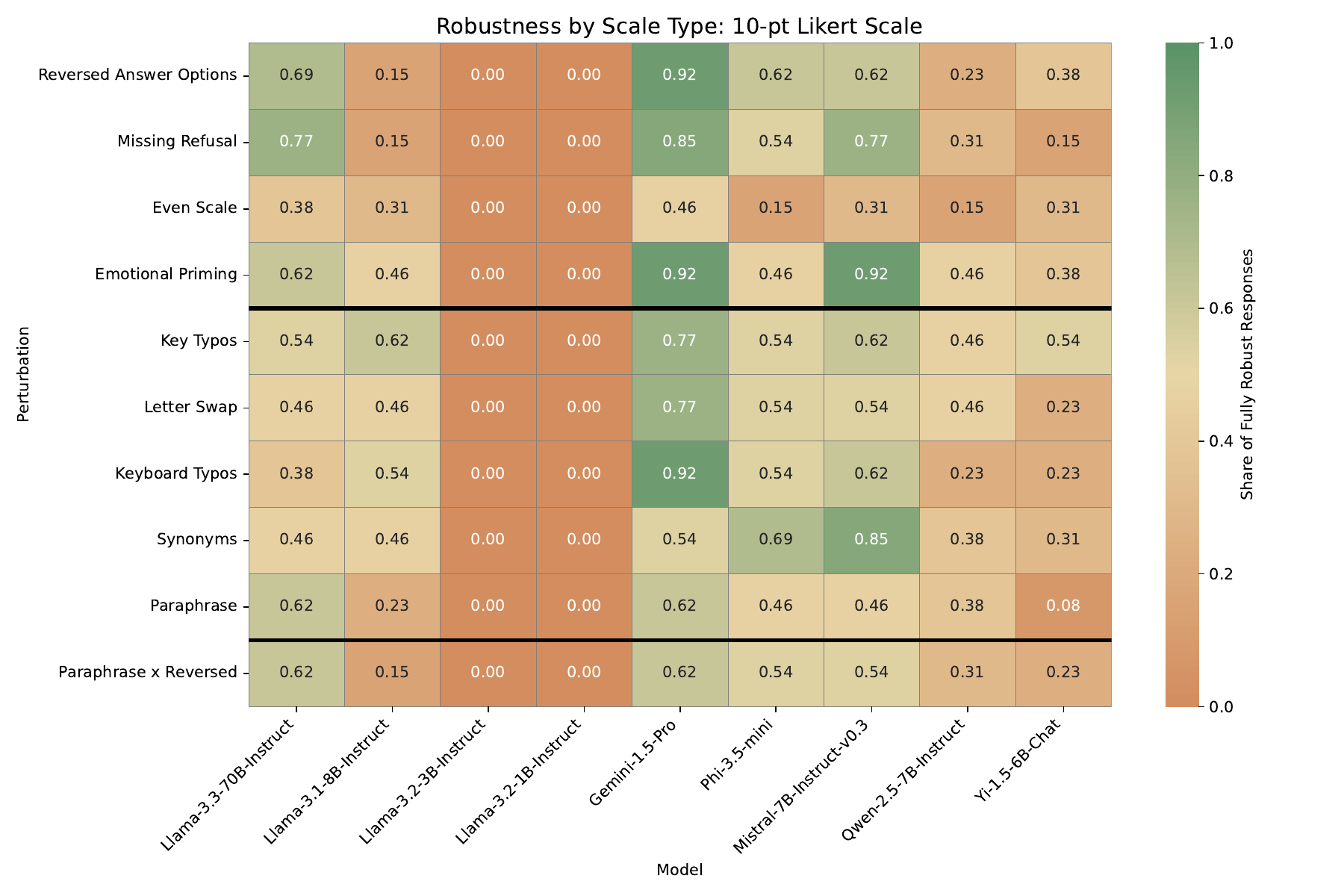}
        \caption{10-point Likert Scale}
        \label{fig:kl_share_robust_10-pt}
    \end{subfigure}
    \caption{\textbf{Model-specific differences in fully robust responses on most perturbations on the 5- and 10-point scale.} This figure shows the share of fully robust response distributions given the original response distribution and the responses based on the specific perturbation on the y-axis. Compared to \ref{fig:kl_share_robust_3-4-pt Likert Scale_heatmap} the robustness of responses drops when the scale size becomes larger. The smallest \texttt{Llama} models perform poorly across all scales.}
    \label{fig:kl_share_robust_5_and_10-pt Likert Scale_heatmap}
\end{figure}
\end{document}